\documentclass[twoside,11pt]{article}

\usepackage[specialissue,accepted]{melba}

\usepackage{mwe} 

\usepackage{amsmath, amsfonts}
\usepackage{bm}

\usepackage{adjustbox}
\usepackage{subcaption}
\usepackage{graphicx}

\melbaid{2024:009}  
\doi{https://doi.org/10.59275/j.melba.2024-f95c}
\melbaauthors{Engelson, Ehrhardt, Kepp, Niemeijer, Handels}  

\volume{2}
\firstpageno{817}  
\melbayear{2024}  
\datesubmitted{01/2024}  
\datepublished{05/2024}  

\melbaspecialissue{MICCAI 2023 LNQ challenge special issue}
\melbaspecialissueeditors{Steve Pieper, Erik Ziegler, Tawa Idris, Bhanusupriya Somarouthu, Reuben Dorent, Gordon Harris, Ron Kikinis}

\ShortHeadings{Lymph Node Segmentation with a Probabilistic Atlas}{Engelson and Ehrhardt}

\title{LNQ Challenge 2023: Learning Mediastinal Lymph Node Segmentation with a Probabilistic Lymph Node Atlas}

\author{\firstname Sofija \surname Engelson\orcid{0009-0007-2493-8107} \thanks{S. Engelson and J. Ehrhardt contributed equally.} \email sofija.engelson@uni-luebeck.de \\  
	\addr Institute of Medical Informatics, University of Lübeck, Lübeck, Germany
	\AND 
	\name Jan Ehrhardt $^*$ \email jan.ehrhardt@uni-luebeck.de \\
	\addr Institute of Medical Informatics, University of Lübeck, Lübeck, Germany \\
    \addr German Research Center for Artificial Intelligence, Lübeck, Germany
    \AND
    \name Timo Kepp\orcid{0000-0003-2024-2958} \email timo.kepp@dfki.de \\
    \addr German Research Center for Artificial Intelligence, Lübeck, Germany
    \AND
    \name Joshua Niemeijer\orcid{0000-0002-2417-8749} \email joshua.niemeijer@dlr.de \\
    \addr German Aerospace Center, Braunschweig, Germany
    \AND
    \name Heinz Handels\orcid{0000-0002-3499-4328} \email heinz.handels@uni-luebeck.de\\
	\addr Institute of Medical Informatics, University of Lübeck, Lübeck, Germany \\
    \addr German Research Center for Artificial Intelligence, Lübeck, Germany
}

\begin{document}

\maketitle

\begin{abstract}
	The evaluation of lymph node metastases plays a crucial role in achieving precise cancer staging, which in turn influences subsequent decisions regarding treatment options. The detection of lymph nodes poses challenges due to the presence of unclear boundaries and the diverse range of sizes and morphological characteristics, making it a resource-intensive process. As part of the LNQ 2023 MICCAI challenge, we propose the use of anatomical priors as a tool to address the challenges that persist in automatic mediastinal lymph node segmentation in combination with the partial annotation of the challenge training data. The model ensemble using all suggested modifications yields a Dice score of 0.6033 and segments 57\% of the ground truth lymph nodes, compared to 27\% when training on CT only. Segmentation accuracy is improved significantly by incorporating a probabilistic lymph node atlas in loss weighting and post-processing. The largest performance gains are achieved by oversampling fully annotated data to account for the partial annotation of the challenge training data, as well as adding additional data augmentation to address the high heterogeneity of the CT images and lymph node appearance.
	Our code is available at~\url{https://github.com/MICAI-IMI-UzL/LNQ2023}.
\end{abstract}



\section{Introduction}
\label{sec:intro}

	In cancer staging, the N-staging component of the Classification of Malignant Tumors (TNM) classification system provides insights into the presence of metastases in regional lymph nodes. Accurate identification of metastatic lymph nodes poses a challenge for diagnosis through CT imaging alone due to minimal contrast differences with surrounding tissue and strong variations in size, shape, number, and location of lymph nodes. PET/CT scans serve as a gold standard to assess functional parameters, that is metabolic activity and, consequently, lymph node malignancy. When PET scans are unavailable due to high examination costs and radioactive exposure to the patient, only CT images are used. In CT, in contrast to PET, only morphological factors such as the lymph node size can be evaluated. The Response Evaluation Criteria in Solid Tumors (RECIST) introduced by \cite{Eisenhauer2009} is commonly used in this context. It defines a lymph node to be pathological if its short-axis diameter exceeds 10\,mm in axial plane. With the above-mentioned challenges of distinguishing lymph nodes from surrounding soft tissue and its highly resource-intensive manual assessment, there arises a need for robust and performant algorithms to tackle the task of detection/segmentation of cancerous lymph node for  lymph node staging without human interaction. Automated segmentation of pathological lymph nodes facilitates tumor staging based on both PET/CT or CT only, and subsequently supports decision-making regarding the necessity of surgery and further treatment.

    The LNQ 2023 challenge hosted at MICCAI 2023 aims to provide a large annotated dataset as well as an organized platform to compare algorithmic methods in the use case of the segmentation of enlarged lymph nodes in the mediastinum. This is specifically relevant for lung cancer patients, but can certainly serve as a benchmark for the extension to other lymph nodes in the human body. The mediastinum contains ten or more lymph nodes, the positioning of which is defined in \cite{El-Sherief2014}.
    
    The problem setting for this challenge consists of three main hurdles that need to be addressed by the participant's approaches:
    First, the provided data in combination with other publicly available datasets is highly heterogeneous regarding the retrieval process and image characteristics such as variances in image resolution and field of view. In addition, the provided challenge data shows images of patients with pathologies (e.g. a collapsed lung, tumors) in the highly individual anatomy of the mediastinum, which makes automatic segmentation and registration particularly difficult. Second, finding lymph nodes is a classic ``looking for a needle in a haystack" problem, as lymph nodes are small. Therefore, the amount of foreground pixels is also small compared to the number of background pixels, which leads to a severe class imbalance. Third, the training dataset for the challenge is weakly annotated, providing segmentation masks of only one or more clinically relevant lymph nodes. The overall results of the challenge show that not accounting for the problem of undersegmentation will result in a significant performance drop on validation and test set.

    This study centers around the use of multiple spatial anatomical priors integrated as supplemental inputs into deep learning methodologies as a tool to address challenges mentioned above. The priors consist of distance maps normalized to the atlas' coordinate system and probability maps indicating the likelihood of lymph node occurrence. To generate the priors, we developed an upstream atlas-to-patient registration approach. The extensive use of additional image augmentation improves generalizability, bridges the domain gap between different datasets, and, in this way, tackles the high data heterogeneity. To address the problem of a high false negative rate due to the strong class imbalance and partial annotation, we use the probabilistic lymph node atlas in loss weighting and post-processing.


\section{Related Works}
\label{sec:related_work}

    The field of automatic lymph node classification, detection, and segmentation has a rich history in medical research. \cite{Feuerstein2012} advanced the generation of an atlas, originally designed for brain imaging, for lymph nodes in the mediastinum to then detect and label the lymph node stations. Similarly, \cite{Feulner2013} developed a probabilistic atlas derived from lymph node segmentation masks, employing it as a spatial prior in a multistep approach based on conventional methods. To refine the prior, the authors smooth the resulting average lymph node segmentations and exclude selected organs, i.e. lungs, the trachea, the esophagus, and the heart.

    The landscape shifted with \cite{Roth2014}, who contributed a dataset of 3D CT volumes from 90 patients with 388 segmented lymph nodes, propelling the rise of learning-based methods in this domain. \cite{Roth2014} trained a convolutional neural network using a 2.5D resampling strategy. As input, the authors use three randomly scaled, rotated and translated orthogonal 2D slices centered on a volume of interest's centroid coordinates. Subsequent researchers, such as \cite{Iuga2021} and \cite{Nayan2022}, expanded on this work by experimenting with diverse network architectures. \cite{Iuga2021} introduced a 3D fully convolutional foveal neural network capable of extracting features at various resolutions, while \cite{Nayan2022} explored a modified upsampling strategy for U-Net++. \cite{oda2018} trained a 3D U-Net to segment lymph nodes as well as other anatomical structures to prevent oversegmentation.

    A promising integration of spatial prior information and learning-based approaches was proposed by \cite{Bouget2021}. The authors incorporated segmentation masks of anatomical structures, such as the esophagus and the azygos vein, as additional channel input to a 3D U-Net. This strategic inclusion aims to prevent the network from generating false positives in these specific areas. In this way, the authors synthesize the strengths of both worlds -- deep learning methods and anatomical priors.


\section{Methods}
\label{sec:methods}

\begin{figure}[h]
    \begin{center}
    \begin{tabular}{c} 
    \includegraphics[width=\textwidth]{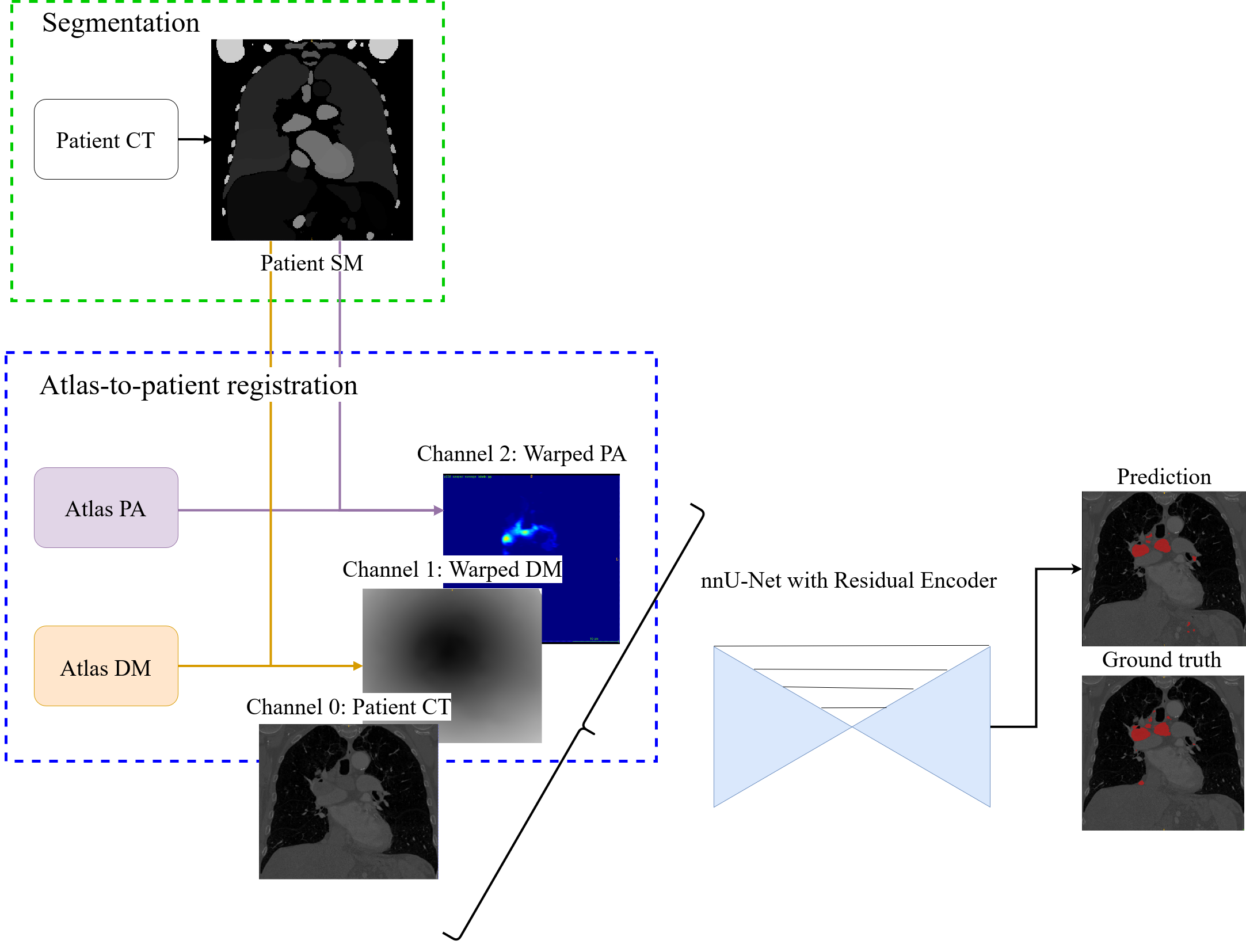}
    \end{tabular}
    \end{center}
    \caption[example] 
    {\label{fig:overview}
    Visualization of the training pipeline.}
    \end{figure}

    Our approach consists of a pre-processing step, the network training and a post-processing of the predicted results. In the pre-processing, automatic segmentation of anatomical structures is followed by atlas-to-patient registration to generate strong anatomical priors, which are used as additional network input, in loss weighting, and in the post-processing step. A U-Net architecture was trained as a segmentation model. The overall framework is shown in Fig.~\ref{fig:overview}.
        
\subsection{Pre-Processing and Generation of Anatomical Priors}
\label{sec:preprocessing}
\begin{figure}[h]
    \begin{subfigure}[t]{\textwidth}
        \centering
        \includegraphics[width=\textwidth]{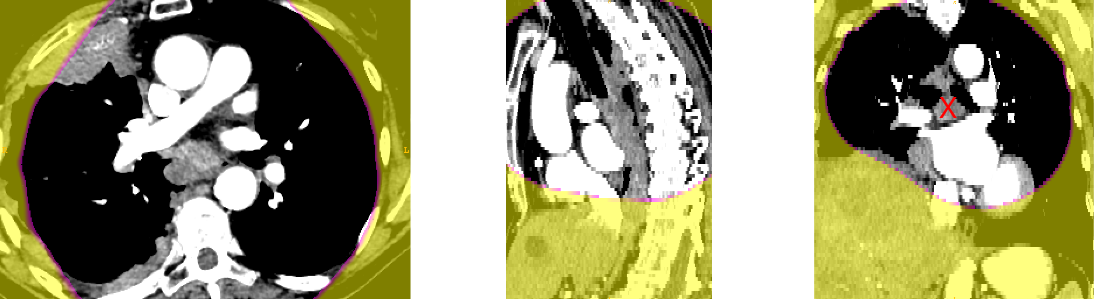}
        \caption{Patient 1.}
    \end{subfigure}
    \begin{subfigure}[t]{\textwidth}
        \centering
        \includegraphics[width=\textwidth]{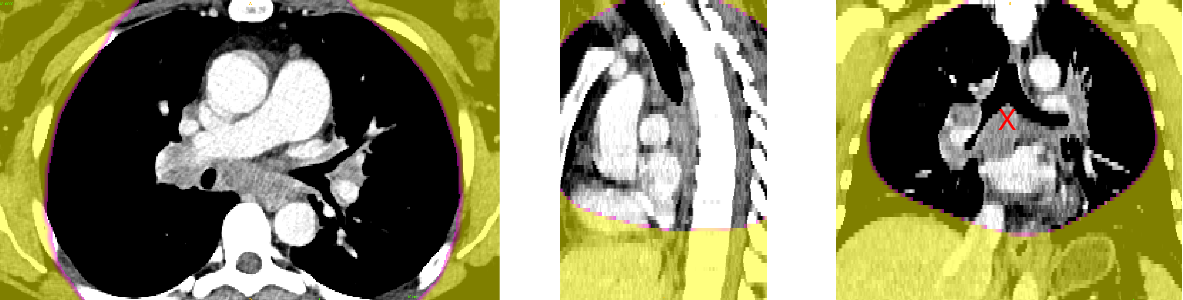}
        \caption{Patient 2.}
    \end{subfigure}
    \caption{Distance maps overlaid over CT for example patients in axial, sagittal, and coronal view. Here, the contrasts of the distance map are set in a way that the contour of the same distance value for both patients is visualized in red. The reference point from which distances are measured is marked with a red cross. The distances are measured in the coordinate system of the atlas patient, thus, the contours show a deformed circle.}
    \label{fig:dm}
    \end{figure}

\begin{figure}[h]
    \begin{subfigure}[t]{\textwidth}
        \centering
        \includegraphics[width=\textwidth]{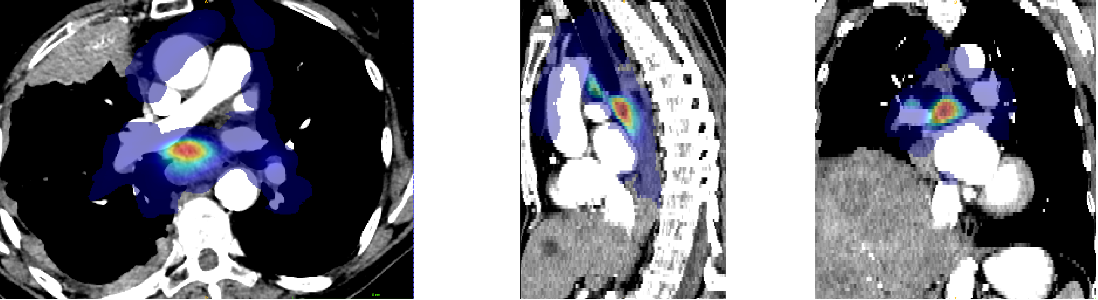}
        \caption{Patient 1.}
        \label{fig:pat1_pa}
    \end{subfigure}
    \begin{subfigure}[t]{\textwidth}
        \centering
        \includegraphics[width=\textwidth]{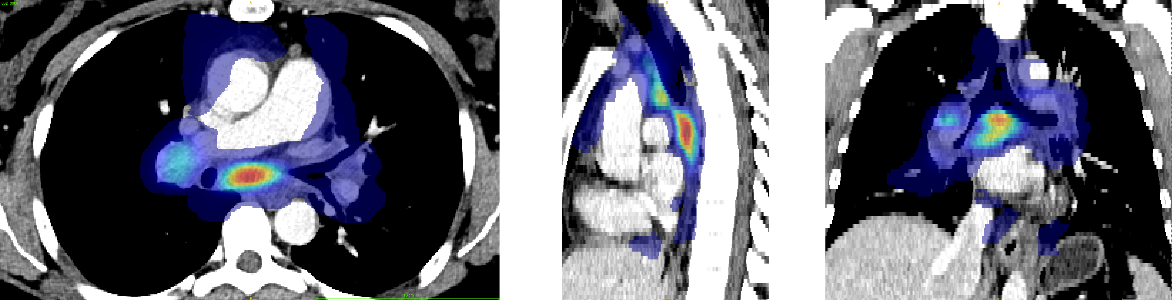}
        \caption{Patient 2.}
        \label{fig:pat2_pa}
    \end{subfigure}
    \caption{Probabilistic lymph node atlases overlaid over CT for example patients in axial, sagittal, and coronal view.}
    \label{fig:pa}
    \end{figure}

    Anatomical prior information was introduced to account for the low contrast of lymph nodes in CT images, strong class imbalance and incomplete labelling of the training data. 
    The proposed prior information consists of anatomical labelling, a distance map (DM) calculated with respect to a specified anatomical landmark (bifurcation of the trachea) defining a kind of anatomical coordinate system, and a probability map for the occurrence of malignant lymph nodes in the training set, referred to as probabilistic lymph node atlas (PA). Visualizations for both priors for example patients are shown on Fig.~\ref{fig:dm} and Fig.~\ref{fig:pa}.
    
    To generate the probabilistic lymph node atlas, all training images were registered to a chest CT lymph node atlas~\citep{Lynch2013} and their ground truth annotations were used to compute occurrence probabilities. The registration pipeline consisted of a sequence of a rigid registration, followed by an affine registration, and finally a deformable registration using a GPU-based implementation of ITK’s VariationalRegistration\footnote{\url{https://itk.org/Doxygen/html/group__VariationalRegistration.html}} module \citep{Werner2014}. The strong heterogeneity of the CT data, as mentioned in Sec.~\ref{sec:intro}, would compromise the robustness of pure intensity-based registration. Therefore, rigid and affine registration were based on segmentation masks (SM) of selected anatomical structures segmented by the TotalSegmentator\footnote{\url{https://github.com/wasserth/totalsegmentator}} algorithm \citep{Wasserthal2023}. The selected structures (i.e. bones, heart, esophagus, trachea, and aorta) include anatomies that are less likely to contain pathologies, which could distort registration results. The resulting occurrence probabilities were smoothed with a Gaussian filter ($\sigma = 5$) and then scaled to a range between zero and one.
    The distance map was also defined in the atlas space with respect to the selected reference point and normalized to a range between zero and one.

    For training and inference, the chest CT atlas was registered to the input images using the registration pipeline described above, and the resulting deformation was used to transfer the probabilistic lymph node atlas and distance map to the subject's coordinate system. Further, the image data was cropped to the bounding box of the lung segmentation masks to reduce computational costs and intensity normalization was performed using nnU-Net's default CT normalization method, which involved taking the 0.5 and 99.5 percentiles of intensity values within the foreground class. This yields values resembling the soft tissue intensity window.

\subsection{Model Architecture and Loss Calculation}
\label{sec:model_arc_loss}
    As a basis for our segmentation network, we used nnU-Net by \cite{Issensee2021}. Inspired by \cite{Isensee2023}, we included additional residual connections in the encoder of the U-Net.
    The loss function was a combination of Dice loss, Cross-Entropy (CE) loss and Tversky loss \citep{Salehi2017}:
    \begin{equation}
        \mathcal{L}(\bm{y},\hat{\bm{y}}) = \lambda_1\mathcal{L}_{\text{CE}}(y_{ijk},\hat{y}_{ijk}) + \lambda_2 \mathcal \sum_{(i,j,k)\in\Omega} w_{ijk}\mathcal{L}_{\text{Dice}}(y_{ijk},\hat{y}_{ijk}) + \lambda_3\mathcal{L}_{\text{Tversky}}(y_{ijk},\hat{y}_{ijk};\alpha, \beta),
    \end{equation}
    where $\Omega$ is the image space, and $\bm{y}$ and $\hat{\bm{y}}$ the ground truth and predicted label maps. The loss weights were chosen to be $\lambda_1 = 0.25$, $\lambda_2 = 0.25$, and $\lambda_3 = 0.5$. The Tversky loss addresses the class imbalance of fore- and background. By adjusting its two hyperparameters, that is, setting $\alpha = 0.25$ and $\beta = 0.75$, we allow a higher false positive rate.

    Additionally, we introduced a loss weighting according to the probabilistic lymph node atlas to further account for the partial annotation of the training data. Due to the partial annotation, false positive predictions can actually be true positives, which are missing in the ground truth annotation. This approach uses a reduced Dice loss weight in areas with high lymph node occurrence. The weight map $W = (w_{ijk})$ at pixel index $(i,j,k)\in\Omega$ was calculated as follows:
    \begin{equation}
        w_{ijk} = \begin{cases} 1 & g_{ijk} = 1 \\  1 - p_{ijk} & g_{ijk} = 0, \ p_{ijk} \leq 0.25 \\ 0.75 & g_{ijk} = 0, \ p_{ijk} > 0.25, \end{cases}
    \end{equation}
    \noindent
    where $G = (g_{ijk}),(i,j,k) \in \Omega$ is the dilated ground truth segmentation with a kernel radius of $2$ and $P = (p_{ijk})$ is the probabilistic lymph node atlas. The ground truth segmentations were dilated to ensure that the lymph node borders are correctly classified. The resulting weight map was multiplied with the pixel-wise Dice loss. This was carried out for multiple resolutions, as loss calculation of the nnU-Net is embedded into a deep supervision framework.

\subsection{Augmentation}
\label{sec:augmentation}
    nnU-Net uses standard augmentations such as random cropping, rotating, and flipping per default. To further enable training with different CT data sets, we adapted the single-source domain generalization technique from \cite{Ouyang2021}. This allows bridging the domain gap caused by different acquisition processes and scanners. During each training iteration, two random intensity or texture transformations were sampled from a shallow convolutional network to generate a global intensity non-linear augmentation (GIN) for each input image. Both GIN augmentations were then merged using interventional pseudo-correlation augmentation (IPA). More precisely, a bias-field-like pseudo-correlation map was used to perform the blending. See the work of \cite{Ouyang2021} for more details. The impact of the GIN and IPA augmentation was controlled by a blending parameter whose weighting follows a Gaussian ramp-up curve \citep{Laine2017} for the first 1,000 epochs of training to gradually introduce domain generalization to the nnU-Net.
  
\subsection{Post-processing}
\label{sec:postprocessing}
    The aim of the proposed post-processing steps is to, on the one hand, ensure that only enlarged lymph nodes are segmented and, on the other hand, account for the partial annotations and strong class imbalance. Consequently, the post-processing consisted of removing connected components that have a diameter smaller than 3, 5 or 7\,mm or skipping this post-processing step. Additionally, we set the threshold for binarization for pixel $ijk$ according to:
    \begin{equation}        
        m_{ijk} = t \times (1 - 0.5 \times p_{ijk}),
        \label{eq:thres_for_bin}
    \end{equation}
    \noindent where $m_{ijk}$ is the threshold depending on the probabilistic lymph node atlas at pixel $ijk$, $t$ is the constant threshold, and $p_{ijk}$ is a pixel at index $i,j,k$ of the probabilistic lymph node atlas $P$. Similarly to \cite{Bouget2021}, the threshold $t$ was kept at $0.5$ or reduced to $0.3$ or $0.2$. This allows for more uncertainty in areas where lymph nodes are more likely to occur and results in enlarged segmentation masks.
    In addition, it was necessary to revert the cropping to the lung segmentation mask by padding the predicted masks to the original input size and removing segmented pixels outside the convex hull of the lung.

    Furthermore, we utilized random variations in network training by ensembling the predictions of five models to account for partial annotation and class imbalance.

\subsection{Semi-Supervised Learning}
\label{sec:SSL}

    The presented approach was trained in a purely supervised manner. However, there is potential to improve generalization by using unsupervised learning techniques. To examine this, we tested several modifications of the semi-supervised learning approach introduced by \cite{Wang2022}. The approach comes from the field of computer vision, and the transferability to the medical image domain has not been researched before. The main idea of the original approach can be summarized as follows: Given two networks with identical network architectures -- a student and a teacher, the student is trained given labeled data as well as unlabeled data with pseudo-labels generated by the teacher. The teacher weights are updated by computing an exponential moving average of the weights of the student. The teacher's task is to differentiate between reliable and unreliable segmented pixels by computing the entropy $H(p_{ijk}) = - \sum_{c=0}^{C-1} p_{ijk}(c) \ log \ p_{ijk}(c)$ of the softmax probabilities $p_{ijk}$ of class $c$ \citep{Wang2022}. A pixel is reliable, if $argmax_c \ p_{ijk}$ exceeds a dynamically increasing threshold, which is computed based on the histogram of entropies. In this way, the teacher generates pseudo-labels for unlabeled data to diversify the training data for the student's learning process.

    The first modification we implemented is a strong augmentation for the student, including GIN and IPA, as well as additional random contrast adjustment and random Gaussian noise. The unlabeled data that was fed to the teacher was not augmented, besides the standard augmentations of nnU-Net. This should ensure the best possible pseudo-label generation by the teacher and a robust segmentation performance through learning on diverse training data by the student.
    
    The downside of the original approach in combination with partially annotated data and the strong class imbalance is that the foreground pixels at the borders of the lymph nodes are often considered to be unreliable. The reason is that the network is more uncertain in detecting lymph nodes than in detecting background. This leads to smaller pseudo-labels, which in turn result in degrading performance. Hence, we developed a second variant that accounts for the problem of undersegmentation. We removed the differentiation between reliable and unreliable pixels and instead use our proposed post-processing approach described in Sec.~\ref{sec:postprocessing} for pseudo-label generation. This strategy artificially enlarges the segmentation masks predicted by the teacher, which in the process of training should lead to the predicted lymph node segmentations to become bigger instead of smaller.


\section{Results}
\label{sec:results}

    The chosen nnU-Net configuration was to train on patches of size 128$\times$112$\times$160 with a batch size of two. 
    The learning rate was set to linearly decrease from 0.01 to $2\times10^{-5}$ until epoch 1,000 and from $2\times10^{-5}$ to $4\times10^{-8}$ until epoch 2,000. This scheduling leads to a more rapid decrease of the learning rate in the first training half than in the second training half. To increase computational efficiency, we implemented a smart cache strategy. More precisely, the items used for training are stored in cache and partially replaced in each iteration.

    The results for supervised training are described in the following sections. For the training of the semi-supervised learning task, we resumed training from the best checkpoint of one of the pre-trained folds. Even though the implementation of the semi-supervised approaches was particularly time-consuming, all proposed implementations led to a degradation of the validation accuracy during training. Typical learning curves can be seen in Fig.~\ref{fig:ssl_loss}.

\subsection{Data}
\label{sec:data}

    We used three different datasets for network training: First, 393 partially annotated CT images from the Mass General Brigham Hospital in Boston are provided as the challenge training dataset \citep{LNQ2023}. Second, a total of 89 patients are part of a dataset collected by the National Institutes of Health in Maryland \citep{Roth2014}. Even though, this dataset contains annotations, they have been refined by \cite{Bouget2021}. The annotations of \cite{Bouget2021} for this dataset as well as another 30 patients from St. Olavs hospital in Trondheim (third dataset) contain lymph nodes of different sizes as well as information about the node's station. For a non-expert, it is difficult to distinguish the segmentation of stations opposed to the segmentation of single lymph nodes with identifiable borders. The inability to properly detect instances and separate collocated lymph nodes as a limitation was addressed and analyzed in \cite{Bouget2021}. The image resolution of the training data varies between $0.58$ and $0.97\,mm^3$ in-plane and $0.5$ to $5.0\,mm^3$ between slices. The field of view can either be limited to the patient's thorax, but can also extend to the whole body. If not marked otherwise, the trained models were tested on the test dataset of the LNQ 2023 challenge. The test set contains 100 patients and a total of 861 segmented lymph nodes in the ground truth annotations. Some additional characteristics of the described datasets can be reviewed in Tab.~\ref{tab:dataset_characteristics}. The lymph node size is the minimum diameter of an ellipse fit to a fully connected component, that is, the segmentation mask of a single lymph node. The minimum ellipsoid diameter can differ slightly from the in-plane diameter as defined in RECIST.

    \begin{table}[h]
        \centering
        \begin{adjustbox}{width=\textwidth}
        \begin{tabular}{lcccc}
        \textbf{Dataset} & \textbf{\# Patients} & \textbf{Total \# LN} & \textbf{AVG \# LN} & \textbf{AVG LN size} \\ \hline
        \cite{Roth2014}   & 89  & 1351 & 15.18 & 8.5482  \\
        \cite{Bouget2021} & 30  & 526  & 17.53 & 6.6453  \\
        LNQ train     & 393 & 556  & 1.41  & 16.5481 \\
        LNQ val       & 20  & 105  & 5.25  & 9.2483  \\
        LNQ test      & 100 & 861  & 8.61  & 8.1831 
        \end{tabular}
        \end{adjustbox}
        \caption{Characteristics of the used datasets.}
        \label{tab:dataset_characteristics}
    \end{table}

\subsection{Metrics}
\label{sec:metrics}
\begin{figure}
        \begin{minipage}[b]{0.6\textwidth}
        \centering
        \includegraphics[width=\textwidth]{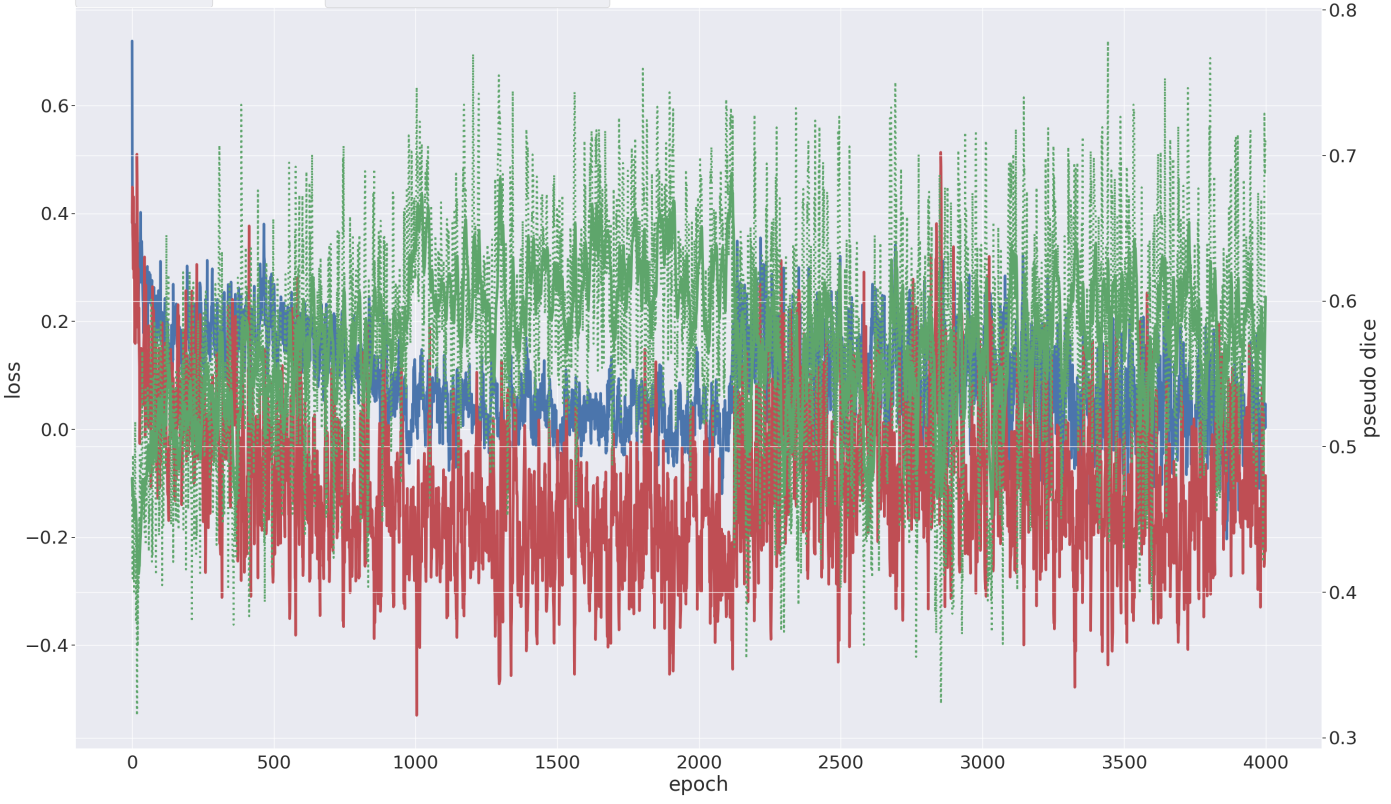}
        \captionof{figure}{Learning curves for a semi-supervised approach. The blue and red curve show the train and the validation loss respectively. The green line is a moving average of the Dice metric. At 2,000 epochs, that is half training time, semi-supervised training starts.}
        \label{fig:ssl_loss}
    \end{minipage}
    \hfill
    \begin{minipage}[b]{0.35\textwidth}
        \centering
        \begin{tabular}{lllll}
        \textbf{Dataset} & \textbf{Dice} & \textbf{ASSD} \\ \hline
        LNQ val       & 0.5600 & 6.7905 \\
        LNQ test      & 0.5690 & 6.8976
        \end{tabular}
        \captionof{table}{Final results of a 5-model ensemble with post-processing on validation and test dataset.}
        \label{tab:final_results}
        \end{minipage}
    \end{figure}

    The evaluation metrics, Dice and Average Symmetric Surface Distance (ASSD), were chosen by the challenge organizers.
    Tab.~\ref{tab:final_results} shows the results of our final submission for the LNQ 2023 challenge for validation and test set. Here, we used an ensemble of five models, three of which were trained with the PA-weighted Dice loss and two were trained without. The threshold for binarization was set to $0.3$ and the minimum diameter of the detected lymph nodes was 5\,mm.

    Additionally, we report Precision and Recall, as well as the percentage of lymph nodes that were correctly found. The metric $\emph{LN found}$ shows the fraction of predicted lymph nodes overlapping with the segmented ground truth lymph nodes, i.e. $\frac{\text{true positives}}{\text{true positives} + \text{false negatives}}$. The calculation of this metric was based on the implementation introduced by the organizers of the BRATS 2023 challenge, initially used for the evaluation of lesion segmentation in brain images.\footnote{\url{https://github.com/rachitsaluja/BraTS-2023-Metrics}} To assess the overlap of the predicted and the ground truth masks, a connected component analysis is carried out based on dilated masks. The dilation factor was set to two. 

\subsection{Ablation Study}
\label{sec:ablation_study}
    In the following, we present an ablation study to quantify the effects of the proposed modifications to the baseline model, which is using the standard nnU-Net trainer with minor modifications to the learning rate and the epoch number on the provided CT data. Models were trained and averaged over five model runs. Training data, network architecture, and training strategy were kept the same for each process. 
    
    The results with and without post-processing are presented in Tab.~\ref{tab:results_nopostprocess} and Tab.~\ref{tab:results_withpostprocess} respectively. The hyperparameters for post-processing were set via grid search. Lowering the threshold for binarization to $0.2$ without any restriction of the minimum lymph node diameter consistently improves the Dice score and Recall. For significance testing, a one-sided t-test (alternative hypothesis: the mean of the distribution underlying the first sample is greater/lower than the mean of the distribution underlying the second sample) has been carried out. The two prior options, that is the additional segmentation masks or the distance map in combination with the probabilistic lymph node atlas, were tested against using only CT as input for training. All other models were compared to the preceding model.

    The use of both prior options leads to an improved Dice score. However, only using SM as a prior yields a p-value smaller than $0.05$ in the one-sided t-test while using DM \& PA as priors is not significant. Surprisingly, a study based on solely fully annotated data showed the opposite effect for using SM as a prior. In this study, the Dice score is approximately three percentage points lower than training on CT only \citep{SPIE2024}. All other suggested configurations improve segmentation performance significantly. A large effect is achieved by accounting for the partial annotation by oversampling the fully annotated datasets, as well as using the Tversky loss to reduce the false negative rate. Tackling the high data heterogeneity by using GIN and IPA as an additional augmentation technique strongly improves performance. The last suggested configuration is using the PA-weighted Dice loss calculation, which, as expected, reduces the amount of false negatives. Generally, the suggested training modifications and post-processing lead to a decreasing false negative rate at the price of an increasing false positives rate. In other words, the Recall improves continuously while maintaining a stable or small decrease in Precision. When training on CT only, around 27\% of the ground truth lymph nodes are found by the model. The combination of all suggested approaches enables the algorithm to find 57\% of the ground truth lymph nodes.
    
    \begin{table}[h]
    \caption{Results averaged over five model runs with different configurations for model training without post-processing. Results significantly different ($p<0.05$) are marked in \emph{italic}. Here, the training with SM and with DM \& PA is compared with training on CT only. The other models are compared with the predecessor model.}
    \label{tab:results_nopostprocess}
    \centering
    \begin{adjustbox}{width=\textwidth}
    \begin{tabular}{cccccc|ccccc}
    \multicolumn{1}{l}{\textbf{DM}} &
      \multicolumn{1}{l}{\textbf{PA}} &
      \multicolumn{1}{l}{\textbf{SM}} &
      \textbf{\begin{tabular}[c]{@{}c@{}}Overs. \\ \& Tver. Loss\end{tabular}} &
      \multicolumn{1}{l}{\textbf{\begin{tabular}[c]{@{}l@{}}GIN \& \\ IPA\end{tabular}}} &
      \textbf{\begin{tabular}[c]{@{}c@{}}PA-weigh.\\ loss\end{tabular}} &
      \textbf{Dice} &
      \textbf{ASSD} &
      \textbf{Precision} &
      \textbf{Recall} &
      \textbf{LN found} \\ \hline
      &   &   &   &   &   & 0.3531 $\pm$ 0.0077 & 14.6567 $\pm$ 0.9442 & 0.8194  & 0.2563 & 0.2652 \\
      &   & X &   &   &   & \emph{0.3818} $\pm$ 0.0127 & \emph{12.8515} $\pm$ 0 5,995 & 0.8107  & 0.2770 & 0.2732 \\
    X & X &   &   &   &   & 0.3611 $\pm$ 0.0138 & 14.1947 $\pm$ 0.8583 & 0.8363  & 0.2684 & 0.2656 \\
    X & X &   & X &   &   & \emph{0.4297} $\pm$ 0.0034 & \emph{11.7228} $\pm$ 0.5704 & 0.8382  & 0.3062 & 0.3104 \\
    X & X &   & X & X &   & \emph{0.5269} $\pm$ 0.0428 & \emph{7.6773} $\pm$ 1.7314  & 0.8061 & 0.4283  & 0.4448 \\
    X & X &   & X & X & X & \emph{0.5509} $\pm$ 0.0053 & \emph{6.9804} $\pm$ 0 3109  & 0.7968 & 0.4590  & 0.4763 \\
    \multicolumn{6}{c|}{ensemble} & 0.5554 & 6.8249  & 0.8148 & 0.4598  & 0.4586
    \end{tabular}
    \end{adjustbox}
    \end{table}

    \begin{table}[h]
    \caption{Results averaged over five model runs with different configurations for model training with post-processing.}
    \label{tab:results_withpostprocess}
    \centering
    \begin{adjustbox}{width=\textwidth}
    \begin{tabular}{cccccc|ccccc}
    \multicolumn{1}{l}{\textbf{DM}} &
      \multicolumn{1}{l}{\textbf{PA}} &
      \multicolumn{1}{l}{\textbf{SM}} &
      \textbf{\begin{tabular}[c]{@{}c@{}}Overs. \\ \& Tver. Loss\end{tabular}} &
      \multicolumn{1}{l}{\textbf{\begin{tabular}[c]{@{}l@{}}GIN \& \\ IPA\end{tabular}}} &
      \textbf{\begin{tabular}[c]{@{}c@{}}PA-weigh.\\ loss\end{tabular}} &
      \textbf{Dice} &
      \textbf{ASSD} &
      \textbf{Precision} &
      \textbf{Recall} &
      \textbf{LN found} \\ \hline
      &   &   &   &   &   & 0.3758 $\pm$ 0.0071 & 14.3196 $\pm$ 0.9879 & 0.7983 & 0.2815 & 0.2749 \\
      &   & X &   &   &   & \emph{0.4024} $\pm$ 0.0122 & \emph{13.0008} $\pm$ 1.0790 & 0.7881 & 0.3020 & 0.2807 \\
    X & X &   &   &   &   & 0.3825 $\pm$ 0.0128 & 14.4255 $\pm$ 1.0931 & 0.8151 & 0.2941 & 0.2754 \\
    X & X &   & X &   &   & \emph{0.4524} $\pm$ 0.0029 & \emph{11.0104} $\pm$.0.6128 & 0.8228 & 0.3344 & 0.3299 \\
    X & X &   & X & X &   & \emph{0.5491} $\pm$ 0.0423 & \emph{7.1048} $\pm$ 1.6649  & 0.7853 & 0.4643  & 0.4896 \\
    X & X &   & X & X & X & \emph{0.5703} $\pm$ 0.006  & \emph{6.4780} $\pm$ 0.3397  & 0.7738 & 0.4943  & 0.5153 \\
    \multicolumn{6}{c|}{ensemble} & \emph{0.6033} &  \emph{5.7483}  & 0.7606 & 0.5483  & 0.5654
    \end{tabular}
    \end{adjustbox}
    \end{table}

    With a Dice score of $0.5690$ the submitted model (Tab.~\ref{tab:final_results}) has a lower performance than the best model with a Dice score of $0.6033$ displayed in Tab.~\ref{tab:results_withpostprocess}. This portrays the following two problems when developing algorithms embedded in a challenge setting: First, the hyperparameter search was not performed systematically, as participants had no access to the fully annotated challenge datasets. Each submission required effort and the number of submissions was limited. And second, hyperparameters for post-processing optimized on the small validation set did not generalize to the test set. As can be reviewed in Tab.~\ref{tab:dataset_characteristics}, the characteristics such as average lymph nodes size differ in train, validation, and test set.


\section{Discussion}
\label{sec:discussion}

    For the task of mediastinal lymph node segmentation, we propose a modified nn-UNet with the use of anatomical priors as additional model input, in loss weighting, and post-processing. The priors are generated with an atlas-to-patient registration approach and serve as orientation reference (distance map) as well as feature guidance (probabilistic lymph node atlas). The ablation study in Sec.~\ref{sec:ablation_study} shows that using anatomical priors provided as additional input alone does not lead to a significant performance gain. However, the probabilistic lymph node atlas enhances loss calculation and post-processing, as well as successfully addresses the challenge of partial annotation and class imbalance. The use of additional (fully annotated) training data and augmentation techniques show the largest effects on the segmentation accuracy. However, using the probabilistic lymph node atlas for post-processing and loss weighting reduces the number of false negatives, which is oftentimes relevant for medical tasks and when only weakly annotated training data is available.

    In the following section, we discuss possible reasons for why the use of the publicly available fully annotated data turned out to be crucially important to yield decent segmentation accuracy. This could also serve as justification for the failed semi-supervised learning approach for the application of mediastinal lymph node segmentation.

    \begin{figure}
        \caption{Probabilities for lymph node occurrence created from (a) the train, (b) the validation, and (c) the test dataset. The probabilities range from 0 to 0.85, the according color map is displayed on the right.}
        \label{fig:comp_probmaps}
         \centering
         \begin{subfigure}[t]{0.28\textwidth}
             \centering
             \includegraphics[width=\textwidth]{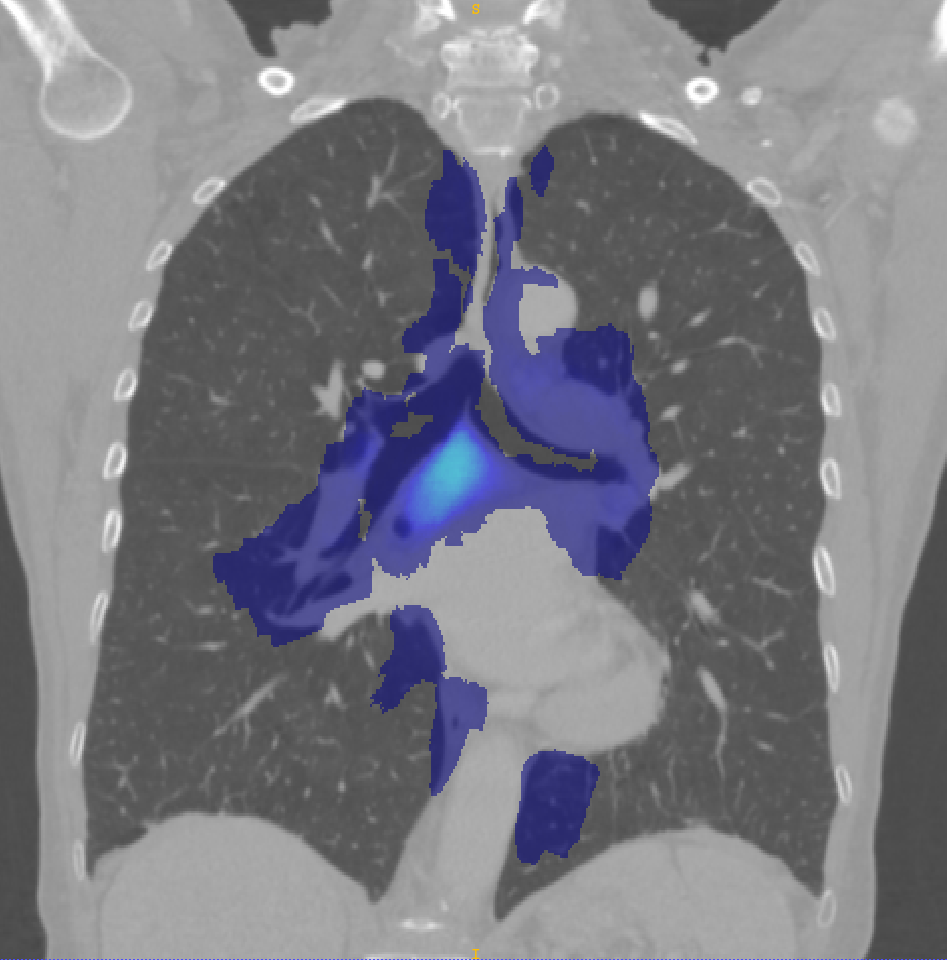}
             \caption{Probabilities for lymph node occurrence in the LNQ 2023 train dataset overlaid on CT of the atlas patient in coronal view.}
             \label{fig:probmap_train}
         \end{subfigure}
         \hfill
         \begin{subfigure}[t]{0.28\textwidth}
             \centering
             \includegraphics[width=\textwidth]{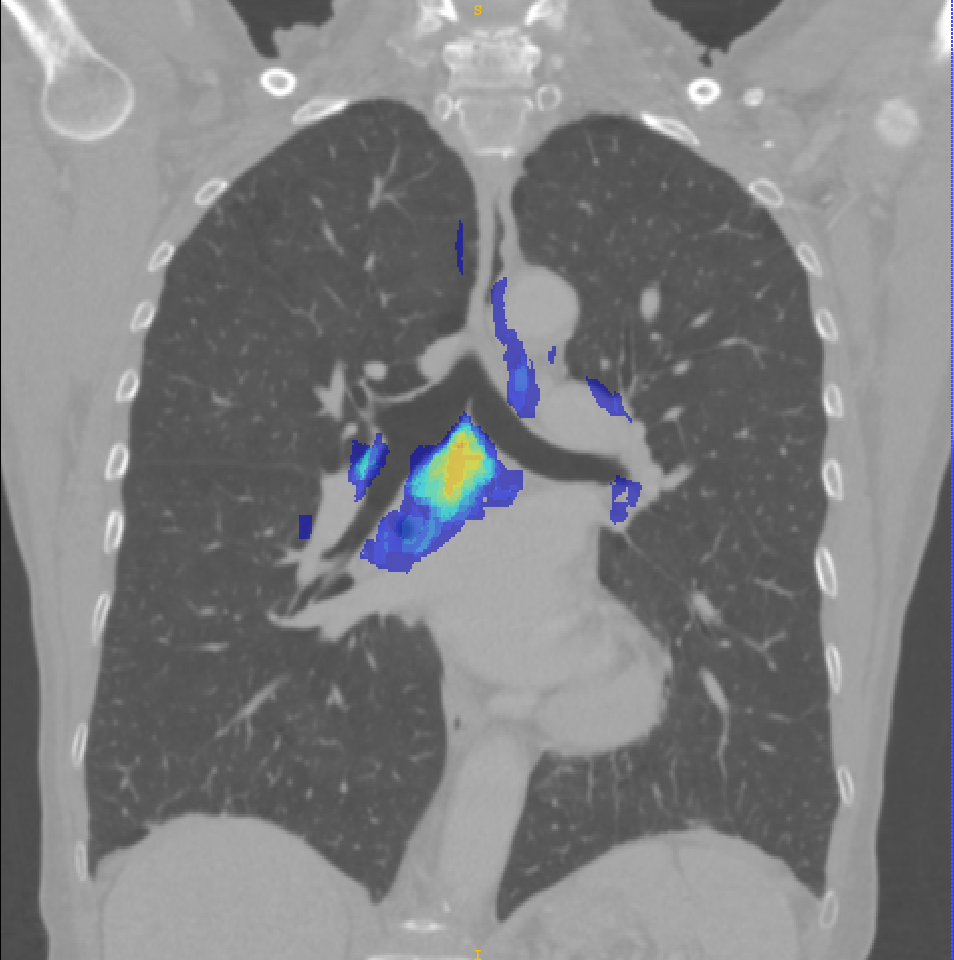}
             \caption{Probabilities for lymph node occurrence in the LNQ 2023 validation dataset overlaid on CT of the atlas patient in coronal view.}
             \label{fig:probmap_val}
         \end{subfigure}
         \hfill
         \begin{subfigure}[t]{0.28\textwidth}
             \centering
             \includegraphics[width=\textwidth]{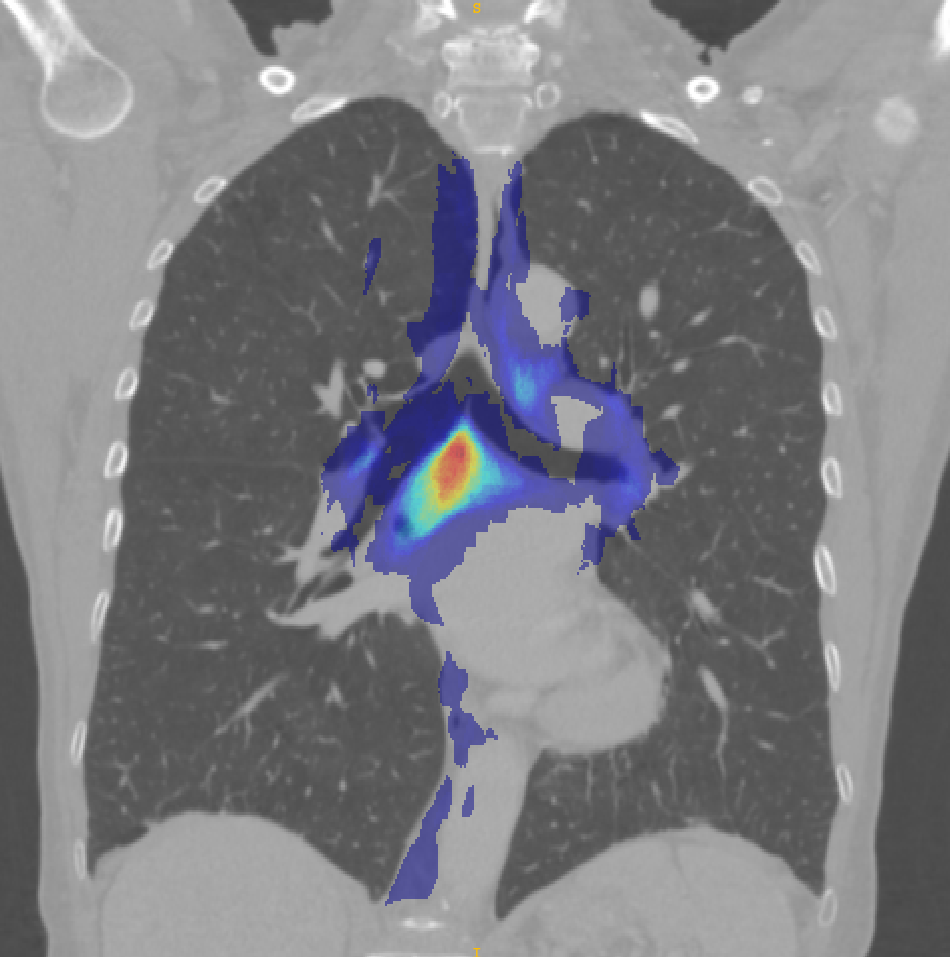}
             \caption{Probabilities for lymph node occurrence in the LNQ 2023 test dataset overlaid on CT of the atlas patient in coronal view.}
             \label{fig:probmap_test}
         \end{subfigure}
         \begin{subfigure}[t]{0.1\textwidth}
             \centering
             \includegraphics[width=0.65\textwidth]{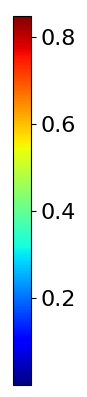}
         \end{subfigure}
    \end{figure}

    Figure~\ref{fig:comp_probmaps} shows the probabilistic lymph node atlases in coronal view generated with the segmentation masks of the LNQ 2023 train dataset in \ref{fig:probmap_train}, the validation dataset in \ref{fig:probmap_val}, and the test dataset in \ref{fig:probmap_test}. According to these images, we find the following differences between train, validation and test dataset:
    \begin{itemize}
        \item The test set includes lymph node stations that are not included in the train set, e.g. between the bottom right lung and the aorta.
        \item Not all stations occur equally often in train, validation and test, e.g. station 7 (beneath the bifurcation of the trachea) and 4L (between aorta and trachea) is segmented more often in val/test than in train.
        \item The size of the segmented masks differs quite heavily, e.g. the segmentation masks in the train set are larger on average. This is confirmed and analyzed in more detail by a co-participant \citep{fischer2024}.
    \end{itemize}
    It can be concluded that the data distribution of the partially annotated train set differs from the fully annotated validation and test set because the choice of which lymph nodes to partially annotate has not been made randomly. On the one hand, it can be expected that medical experts are more likely to annotate strongly enlarged lymph nodes instead of lymph nodes that are pathological, but smaller. On the other hand, some stations are easier to find in the mediastinum than others and, therefore, are examined more often, e.g. the location of station 7 is distinct for most patients and pathologies.
    Self-enhancing methods, such as the proposed semi-supervised learning approach, reproduce results learned the by the model trained in a supervised manner on the data distribution of the training data. If the data or annotations of the training dataset carry biases, they get enforced with further unsupervised training. More extensive hyperparameter tuning could have possibly improved results by tackling the problem of a high false negative rate, or in other words "the problem of not segmenting enough". However, this, most probably, would have not mitigated the consequences of the biased partial annotation, i.e. "the problem of not segmenting the right thing".

    The incorporation of anatomical priors, or in some way encoded prior knowledge, into the learning process of deep learning methods does seem intuitive when dealing with tasks of high complexity and label uncertainty, such as in the task presented in the LNQ 2023 challenge. It is necessary to point out that the anatomical priors were only provided as additional inputs to the algorithm, without controlling whether the algorithm learns features from them. Possibly, CT already contains all necessary information to segment pathological lymph nodes. Simultaneously, the segmentation performance does have potential for improvement -- the best performing model still misses to segment over 40\% of the pathological lymph nodes. In medical image applications, oftentimes, training datasets are small, but the task complexity is high. Therefore, algorithms are particularly at risk of showing the "Clever Hans" behavior, that is predictors making correct predictions for the wrong reasons \citep{Lapushkin20219}. Examples of this are shown in \cite{degrave_ai_2021}, \cite{badgeley_deep_2019}, and \cite{zech_variable_2018}. Also, in the application of mediastinal lymph node segmentation, it is difficult to say, whether the features encoded in the U-Net encoder are semantically meaningful. The symbiosis of prior knowledge with the learning process has the potential to ensure that the learned features actually are responsible for class changes. Therefore, it is necessary to explore other options of incorporating prior knowledge in the network learning process in further research. To further improve segmentation accuracy for the underlying use case, it might be interesting to use the fully annotated validation and test data for training.


\acks{This work was supported by grants of the collaborative research project ``AI ecosystem in health care" within the subproject titled ``Health system-based analysis of disease patterns using lung diseases as an example'' financed by the federate state Schleswig-Holstein, Germany.}

\clearpage

%
\ethics{The work follows appropriate ethical standards in conducting research and writing the manuscript, following all applicable laws and regulations regarding treatment of animals or human subjects.}

\coi{We declare we don't have conflicts of interest.}


\bibliography{sources}


\clearpage
\appendix

\section{Generation of the Probabilistic Lymph Node Atlas}
\label{app:generation_prob_atlas}

\begin{figure}[h]
\begin{center}
\begin{tabular}{c} 
\includegraphics[width=\textwidth]{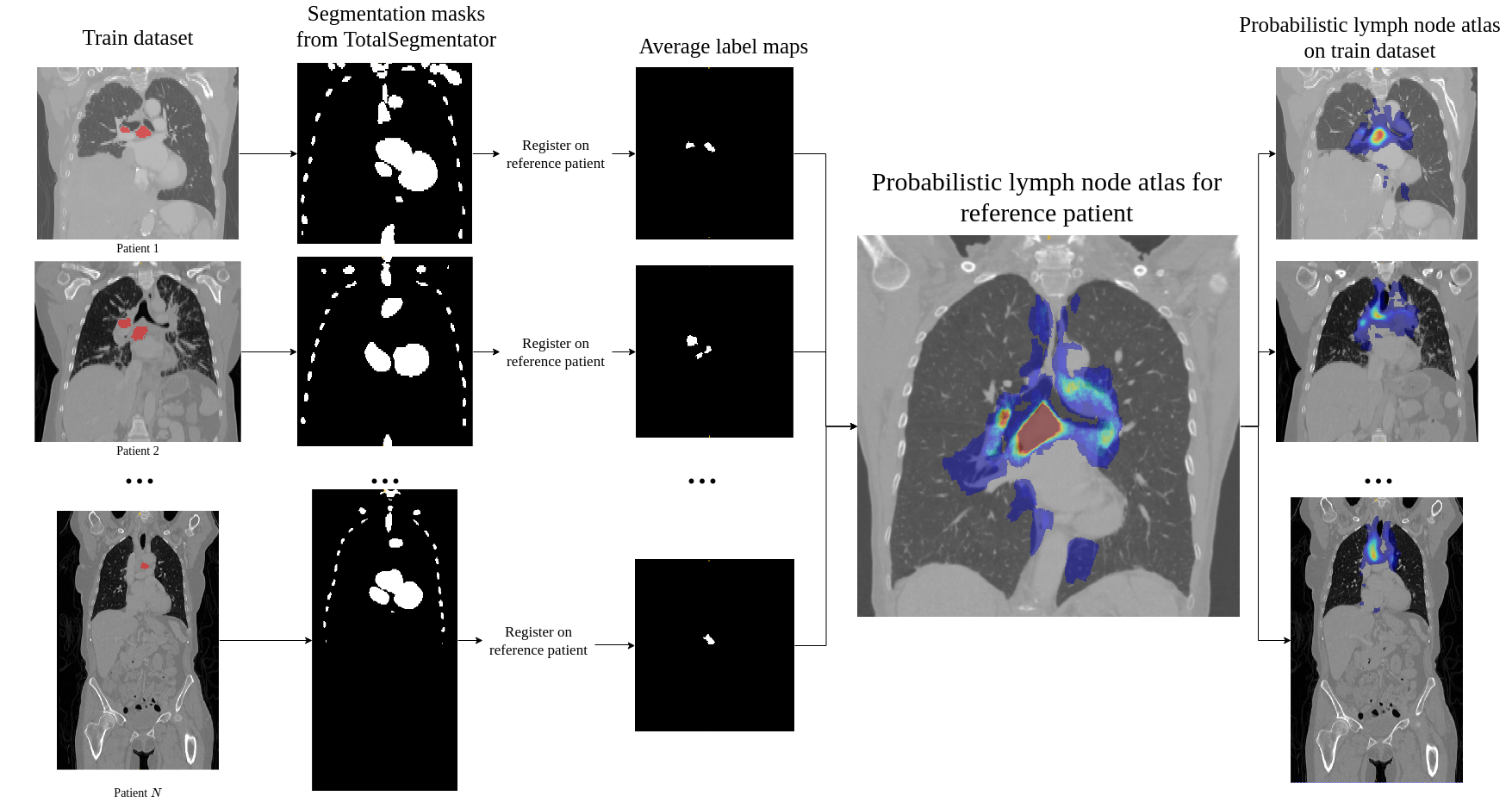}
\end{tabular}
\end{center}
\caption[example] 
{\label{fig:atlas_gen}
The probabilistic lymph node atlas originates from the registration of annotated, publicly available CT images of 512 patients to an atlas image \citep{Roth2014, Bouget2021, LNQ2023}. The CT of patient two from the data provided by \cite{Lynch2013} serves as an atlas. The resulting displacement fields are used to warp the segmentation masks of the training data to the atlas. The registered, segmented lymph nodes are averaged to create a probability map. As a last step, the probabilistic lymph node atlas is registered back to the training data.}
\end{figure}

\end{document}